\documentclass[11pt,a4paper]{article}
\usepackage{times,latexsym}
\usepackage{url}
\usepackage[T1]{fontenc}
\usepackage{graphics}
\usepackage{subcaption}
\usepackage{tikz}
\usepackage{tikz-qtree}
\usepackage{bm}
\usepackage{tikz-dependency}
\usepackage[acceptedWithA]{tacl2018v2}

\usepackage{xcolor}

\usepackage{times}
\usepackage{amsmath}
\usepackage{amssymb}
\usepackage{latexsym}
\usepackage{bm}
\usepackage{booktabs}
\usepackage{footnote}
\usepackage{tablefootnote}
\usepackage{cleveref}
\usepackage{todonotes}
\usepackage{commath}
\usepackage{algorithm}
\usepackage{xytree}
\usepackage{tikz-dependency}
\usepackage{multirow}

\usepackage{algpseudocode}
\makeatletter
\newcommand*{\textoverline}[1]{$\overline{\hbox{#1}}\m@th$}
\makeatother
\usepackage{tikz}

\newcommand*\bluecircled[1]{\tikz[baseline=(char.base)]{%
            \node[shape=circle,fill=blue!20,draw,inner sep=2pt, minimum size=17.5pt] (char) {#1};}}

\newcommand*\yellowcircled[1]{\tikz[baseline=(char.base)]{%
            \node[shape=circle,fill=yellow!20,draw,inner sep=2pt, minimum size=17.5pt] (char) {#1};}}
\newcommand*\redcircled[1]{\tikz[baseline=(char.base)]{%
            \node[shape=circle,fill=red!20,draw,inner sep=0pt, minimum size=17.5pt] (char) {#1};}}
\usetikzlibrary{bayesnet}

\usepackage{microtype}
\usepackage{pifont}
\newcommand{\cmark}{\ding{51}}%
\newcommand{\xmark}{\ding{55}}%

\usepackage{xspace,mfirstuc,tabulary}

\newif\iftaclinstructions
\taclinstructionsfalse 
\iftaclinstructions
\renewcommand{\confidential}{}
\renewcommand{\anonsubtext}{(No author info supplied here, for consistency with
TACL-submission anonymization requirements)}
\newcommand{\instr}
\fi

\iftaclpubformat 

\else

\fi

\title{The Return of Lexical Dependencies:\\ Neural Lexicalized PCFGs}

\author{
 Hao Zhu, Yonatan Bisk, Graham Neubig \\
 Language Technologies Institute, Carnegie Mellon University \\
  {\tt zhuhao@cmu.edu, \{ybisk, gneubig\}@cs.cmu.edu } \\
}

\date{}

\begin{document}
\maketitle
\begin{abstract}
In this paper we demonstrate that \textit{context free grammar (CFG) based methods for grammar induction benefit from modeling lexical dependencies}. This contrasts to the most popular current methods for grammar induction, which focus on discovering \textit{either} constituents \textit{or} dependencies. Previous approaches to marry these two disparate syntactic formalisms (e.g. lexicalized PCFGs) have been plagued by sparsity, making them unsuitable for unsupervised grammar induction.  However, in this work, we present novel neural models of lexicalized PCFGs which allow us to overcome sparsity problems and effectively induce both constituents and dependencies within a single model. Experiments demonstrate that this unified framework results in stronger results on both representations than achieved when modeling either formalism alone.\footnote{Code is available at \url{https://github.com/neulab/neural-lpcfg}.}

\end{abstract}

\section{Introduction}


Unsupervised grammar induction aims at building a formal device for discovering syntactic structure from natural language corpora. 
Within the scope of grammar induction, there are two main directions of research: unsupervised constituency parsing, which attempts to discover the underlying structure of phrases, and unsupervised dependency parsing, which attempts to discover the underlying relations between words. Early work on induction of syntactic structure focused on learning phrase structure and generally used some variant of probabilistic context-free grammars (PCFGs; \citet{lari1990estimation,charniak1996tree,clark2001unsupervised}). In recent years, dependency grammars have gained favor as an alternative syntactic formulation \cite{yuret1998discovery,carroll1992two,paskin2002grammatical}. Specifically, the dependency model with valence (DMV) \cite{klein2004corpus} forms the basis for many modern approaches in dependency induction. Most recent models for grammar induction, be they for PCFGs, DMVs, or other formulations, have generally coupled these models with some variety of neural model to use embeddings to capture word similarities, improve the flexibility of model parameterization, or both \cite{he2018unsupervised,jin2019unsupervised,kim2019compound,han2019enhancing}.

\begin{figure}[!t]
\small
\begin{tikzpicture}[level distance=25pt]
\Tree [.S[{\sc chasing}] [.NP[{\sc dog}] [.DT[{\sc the}] the ] [.NN[{\sc dog}] dog ]] [.VP[{\sc chasing}] [.VBZ[{\sc is}] is ] [.VP[{\sc chasing}] \edge[roof]; {chasing the cat} ]]]
\end{tikzpicture}
\caption{Lexicalized phrase structure tree for ``the dog is chasing the cat.'' The head word of each constituent is indicated with parentheses.}
\label{fig:lexicalized-phrase-structure-tree}
\end{figure}
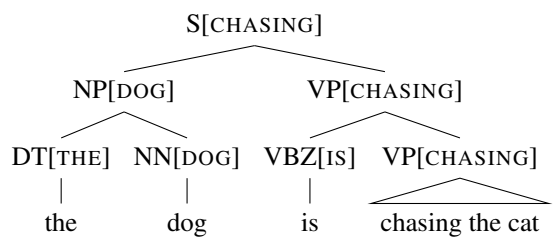

Notably, the two different syntactic formalisms capture very different views of syntax.
Phrase structure takes advantage of an abstracted recursive view of language, while the dependency structure more concretely focuses on the propensity of particular words in a sentence to relate to each-other syntactically.
However, few attempts at unsupervised grammar induction have been made to marry the two and let both benefit each other.
This is precisely the issue we attempt to tackle in this paper.

As a specific formalism that allows us to model both formalisms at once, we turn to lexicalized probabilistic context-free grammars (L-PCFGs; \citet{collins2003head}).
L-PCFGs borrow the underlying machinery from PCFGs but expand the grammar by allowing rules to include information about the lexical heads of each phrase, an example of which is shown in \Cref{fig:lexicalized-phrase-structure-tree}.
The head annotation in the L-PCFG provides lexical dependencies that can be informative in estimating the probabilities of generation rules.
For example, the probability of VP[{\sc chasing}] $\rightarrow$ VBZ[{\sc is}] VP[{\sc chasing}] is much higher than VP $\rightarrow$ VBZ VP, because ``chasing'' is a present participle
.
Historically, these grammars have been mostly used for \emph{supervised} parsing, combined with traditional \emph{count-based} estimators of rule probabilities \cite{collins2003head}.
Within this context, lexicalized grammar rules are powerful, but the counts available are sparse, and thus required extensive smoothing 
to achieve competitive results \cite{bikel2004intricacies,Hockenmaier2002}.



\begin{figure}[!t]
    \centering
    \includegraphics[width=\linewidth]{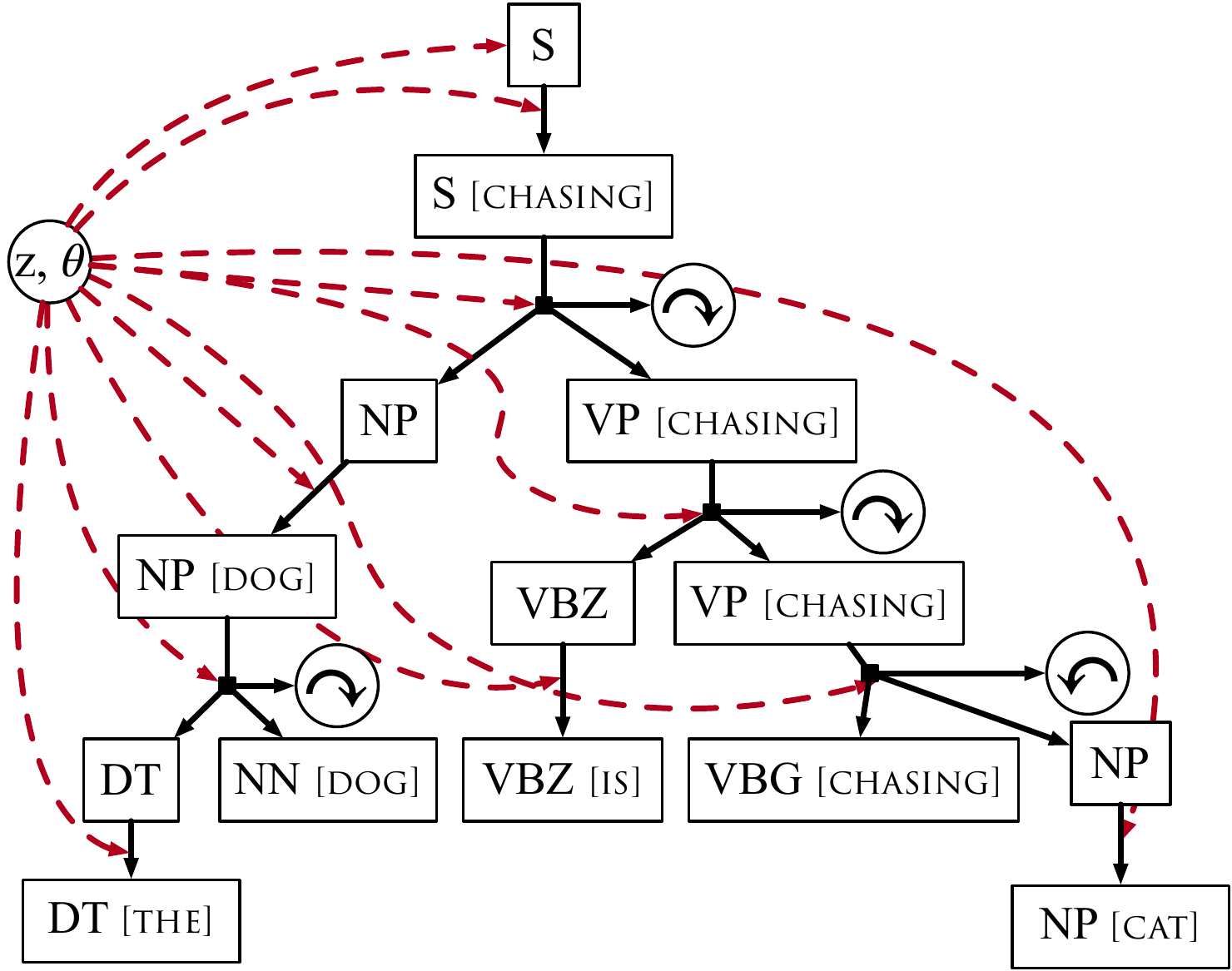} 
    \caption{Model Diagram of Lexicalized Compound PCFG. 
    Black lines indicate production rules, and red dashed lines indicate that the compound variable and parameters participating in productions.}
    \label{fig:diagram}
\end{figure}

In this paper, we contend that with recent advances in neural modeling, it is time to \textbf{return to modeling lexical dependencies}, specifically in the context of unsupervised constituent-based grammar induction.
We propose neural L-PCFGs as a parameter-sharing method to alleviate the sparsity problem of lexicalized PCFGs. \Cref{fig:diagram} illustrates the generation procedure of a neural L-PCFG.
Different from traditional lexicalized PCFGs, the probabilities of production rules are not independently parameterized, but rather conditioned on the representations of non-terminals, preterminals and lexical items (\S\ref{sec:scoring-function}).
Apart from devising lexicalized production rules (\S\ref{sec:lexicalized-context-free-grammar}) and their corresponding scoring function, we also follow \citet{kim2019compound}'s compound PCFG model for (non-lexicalized) constituency parsing with compound variables (\S\ref{sec:compound_grammar}), enabling modeling of a continuous mixture of grammar rules.%
\footnote{In other words, we do not induce a single PCFG, but a distribution over a family of PCFGs.}
We define how to efficiently train (\S\ref{sec:training}) and perform inference (\S\ref{sec:inference}) in this model using dynamic programming and variational inference.

Put together, we expect this to result in a model that both is effective, and \emph{simultaneously} induces both phrase structure and lexical dependencies,%
\footnote{Note that by ``lexical dependencies'' we are referring to unilexical dependencies between the head word and child non-terminals, as opposed to bilexical dependencies between two words (as are modeled in many dependency parsing models).}
whereas previous work has focused on only one.
Our empirical evaluation examines this hypothesis, asking the following question:

{\em\noindent\begin{quote}
   In neural grammar induction models, is it possible to jointly and effectively learn both phrase structure and lexical dependencies? Is using both in concert better at the respective tasks than specialized methods that model only one at a time?\end{quote}
}

Our experiments (\S\ref{sec:effectiveness}) answer in the affirmative, with 
better performance than baselines designed specially for either dependency or constituency parsing under multiple settings.
Importantly, our detailed ablations show that 
methods of factorization play important role in the performance of neural L-PCFGs (\S\ref{sec:ablation_study}). Finally, qualitatively (\S\ref{sec:qualitative_analysis}), we find that latent labels induced by our model align with annotated gold non-terminals in PTB. 


\section{Motivation and Definitions}
In this section, we will first provide the background of constituency grammars and dependency grammars, and then formally define the general L-PCFG, illustrating how both dependencies and phrase structures can be induced from L-PCFGs.

\subsection{Phrase Structures and CFGs}
\begin{center}
\def\qtreeunaryht{1.5ex}
\def\qroofy{1}
\def\qroofx{4}
\def\qroofpadding{0.1em}
\begin{tikzpicture}[level distance=20pt]
\Tree [.S [.NP [.DT the ] [.NN dog ]] [.VP [.VBZ is ] [.VP \edge[roof]; {chasing the cat} ]]]
\end{tikzpicture}
\end{center}



The phrase structure of a sentence is formed by recursively splitting constituents. In the parse above: the sentence is split into a noun phrase (NP) and a verb phrase (VP), which can themselves be further split into smaller constituents; for example, the NP is comprised of a determiner (DT) ``the'' and a normal noun (NN) ``dog.'' 

Such phrase structures are represented as a context-free grammar\footnote{Note $\epsilon \notin \mathcal{T}$ and $\Sigma \cap \mathcal{T} = \varnothing$, so this formulation does not capture the structure of sentences of length zero or one.}  (CFG), which can generate an infinite set of sentences via the repeated application of a finite set $\mathcal{R}$ of rules:
\begin{alignat*}{3}
        S &\rightarrow A, &&A \in \mathcal{N}\\
        A &\rightarrow BC, \quad&&A \in \mathcal{N}, B, C \in \mathcal{N}\cup\mathcal{P}\\
        T &\rightarrow \alpha, &&T \in \mathcal{P}
\end{alignat*}
$S$ denotes a start symbol, $\mathcal{N}$ is a finite set of non-terminals, $\mathcal{P}$ is a finite set of preterminals, $\Sigma$ is a set of terminal symbols, i.e. words and punctuation. 

\subsection{Dependency Structures and Grammars}
\begin{center}
\begin{dependency}[theme = simple]
   \begin{deptext}[column sep=1em]
      The \& dog \& is \& chasing \& the \& cat \\
   \end{deptext}
   \deproot[edge height=1.0cm]{4}{ROOT}
   \depedge{2}{1}{det}
   \depedge{4}{2}{nsubj}
   \depedge{4}{3}{aux}
   \depedge{4}{6}{nobj}
   \depedge{6}{5}{det}
\end{dependency}
\end{center}

In a dependency tree of a sentence, the syntactic nodes are the words in the sentence. Here the root is the \textbf{root word} of the sentence, and the children of each word are its \textbf{dependents}. Above, the root word is \textit{chasing}, which has three dependents, its subject (nsubj) \textit{dog}, auxiliary verb (aux) \textit{is}, and object (nobj) \textit{cat}. 
A dependency grammar\footnote{This work assumes a projective tree.
} 
specifies the possible \textbf{head}-\textbf{dependent} pairs $\mathcal{D} = \{(\alpha_i, \beta_i)\}\in(\mathcal{V}\cup\{\text{ROOT}\})\times\mathcal{V}$, where the set $\mathcal{V}$ denotes the vocabulary.

\subsection{Lexicalized CFGs}
\label{sec:lexicalized-context-free-grammar}

Although both the constituency and dependency grammars capture some aspects of syntax, we aim to leverage their relative strengths in a single unified formalism. 
In a unified grammar, these two types of structure can benefit each other. 
For example, in \textit{The dog is chasing the cat of my neighbor's}, while the phrase \textit{of my neighbor's} might be incorrectly marked as the adverbial phrase of \textit{chasing} in a dependency model, the constituency parser can provide the constraint that \textit{the cat of my neighbor's} is a constituent, thereby requiring \textit{chasing} to be the head of the phrase. 

Lexicalized CFGs are based on a backbone similar to standard CFGs but parameterized to be 
sensitive to lexical dependencies such as those used in dependency grammars.
Similarly to CFGs, L-CFGs are defined as a five-tuple  $\mathcal{T} = (S,\mathcal{N},\mathcal{P},\Sigma,\mathcal{R})$.
The differences lie in the formulation of rules $\mathcal{R}$: 
\begin{alignat*}{3}
        \text{\bluecircled{1} } && S &\rightarrow A[\alpha], &&A \in \mathcal{N}\\
        \text{\yellowcircled{2l} }&&A[\alpha] &\rightarrow B[\alpha] C[\beta], \quad&&A \in \mathcal{N}, B, C \in \mathcal{N}\cup\mathcal{P}\\
        \text{\yellowcircled{2r} }&&A[\alpha] &\rightarrow B[\beta] C[\alpha], &&A \in \mathcal{N}, B, C \in \mathcal{N}\cup\mathcal{P}\\
        \text{\redcircled{3} }&& T[\alpha] &\rightarrow \alpha, &&T \in \mathcal{P}
\end{alignat*}
where $\alpha, \beta \in \Sigma$ are words, and mark the head of constituent when they appear in ``$[\cdot]$''.\footnote{w.l.o.g. we only consider binary branching in $\mathcal{T}$.} Branching rules \yellowcircled{2l} and \yellowcircled{2r} encode the dependencies $(\alpha, \beta)$.\footnote{Note that root seeking rule \bluecircled{1} encodes $(\text{ROOT}, \alpha)$.}

In a lexicalized CFG, a sentence $\bm{x}$ can be generated by iterative binary splitting and emission, forming a \textbf{parse tree} $t=[r_1, r_2, \dots, r_{2|\bm{x}|}]$, where rules $r_i$ are sorted from top to bottom and from left to right. We will denote the set of parse trees that generate $\bm{x}$ within grammar $\mathcal{T}$ as $\mathcal{T}_{\bm{x}}$.

\subsection{Grammar Induction with L-PCFGs}
In this subsection, we will introduce L-PCFGs, the probabilistic formulation for L-CFGs. The task of \emph{grammar induction} is to ask, given a corpus $\mathcal{C} \subset \Sigma^{+}$, how can we obtain the probabilistic generative grammar that maximizes its likelihood.
With the induced grammar, we are also interested in how to obtain the trees that are most likely given an individual sentence -- in other words, syntactic parsing according to this grammar.

We begin by defining the probability distribution over sentences $\bm{x}$, by marginalizing over all parse trees that may have generated $\bm{x}$:

\begin{equation}
    p_{\bm{z}}(\bm{x}) = \sum_{t \in \mathcal{T}_{\bm{x}}} p_{\bm{z}}(t) = \frac{1}{\mathcal{Z}(\mathcal{T}, \bm{z})}\sum_{t\in\mathcal{T}_{\bm{x}}} \tilde{p}_{\bm{z}}(t),
\end{equation}
where $\tilde{p}_{\bm{z}}(t)$ is an unnormalized probability of a parse tree (which we will refer to as an \textit{energy function}), $\mathcal{Z}(\mathcal{T}, \bm{z}) = \sum_{t\in\mathcal{T}} \tilde{p}_{\bm{z}}(t)$ is the normalizing constant, and $\bm{z}$ is a compound variable (\S\ref{sec:compound_grammar}) which allows for more complex and expressive generative grammars \cite{robbins1951asymptotically}. 

We define the energy function of a parse tree by exponentiating a score $G_{\theta}(t, \bm{z})$
\begin{equation}
\tilde{p}_{\bm{z}}(t) \propto \exp G_{\theta}(t, \bm{z})
\end{equation}
where $\theta$ is the parameter of function $G_{\theta}$.
Theoretically, $G_{\theta}(t)$ could be an arbitrary scoring function, but in this paper, as with most previous work,
we consider a context-free scoring function, where the score of each rule $r_i$ is independent of the other rules in the parse tree $t$:
\begin{equation}
    G_{\theta}(t, \bm{z}) = \sum_{i = 1}^k g_{\theta}(r_i, \bm{z}),
\end{equation}
where $g_{\theta}(r, \bm{z}): \mathcal{R} \times \mathbb{R}^n \rightarrow \mathbb{R}$ is the rule-scoring function which maps the rule and latent variable $z\in \mathbb{R}^n$ to real space, assigning a log likelihood to each rule.
This formulation allows for efficient calculation using dynamic programming.
We also include a restriction that the energies must be \emph{top-down locally-normalized},
under which the partition function should automatically equate to 1
\begin{equation}
\mathcal{Z}(\mathcal{T}, \bm{z}) = \sum_{t\in\mathcal{T}} \exp G_{\theta}(t, \bm{z}) = 1
\end{equation}

To train an L-PCFG, we maximize the log likelihood of the corpus (the latent variable is marginalized out):
\begin{equation}
    \theta^* = \arg\max_{\theta} \sum_{\bm{x}\in\mathcal{C}} \log \mathbb{E}_{\bm{z}} p_{\bm{z}}(\bm{x})
\end{equation}

And obtain the most likely parse tree of a sentence by maximizing the posterior probability:
\begin{equation}
    t^* = \arg\max_{t\in\mathcal{T}_{\bm{x}}} \mathbb{E}_{\bm{z}} \exp G_{\theta^*}(t, \bm{z})
\end{equation}

\section{Neural Lexicalized PCFGs}
\label{sec:scoring-function}


As noted, one advantage of L-PCFGs is that the obtained $t^*$ encodes both dependencies and phrase structures, allowing both to be induced simultaneously.
We also expect this to improve performance, because different information is capture by each of these two structures. 
However, this expressivity comes at a price: more complex rules.
In contrast to the traditional PCFG, which has $\mathcal{O}(|\mathcal{N}|(|\mathcal{N}| + |\mathcal{P}|)^2)$ production rules, the L-PCFG requires $\mathcal{O}(|V||\mathcal{N}|(|\mathcal{N}| + |\mathcal{P}|)^2)$ production rules.
Because traditionally rules of L-PCFGs have been parameterized independently by scalars, i.e. $g_{\theta}(r_i, \bm{z}) = \theta_i$ \citep{collins2003head}, these parameters were hard to estimate due to data sparsity.

We propose an alternate parameterization, the \emph{neural L-PCFG}, which ameliorates these sparsity problems through parameter sharing, and the \emph{compound L-PCFG}, which allows a more flexible sentence-by-sentence parameterization of the model. Below, we explain the neural L-PCFG factorization we found performed best but include ablations of our decisions in \Cref{sec:impact_of_factorization}.



 

\subsection{Neural L-PCFG Factorization}
The score of an individual rule is calculated as the combination of several component probabilities:
\begin{description}
\item[root to non-terminal probability $p_{\bm{z}}(S\rightarrow A)$:] Probability that the start symbol produces a non-terminal $A$.\footnote{i.e. $A$ is the non-terminal of the whole sentence}
\item[word emission probability $p_{\bm{z}}(A\rightarrow \alpha)$:] Probability that the head word of a constituent is $\alpha$ conditioned on that the non-terminal of the constituent is $A$.
\item[{\parbox[t]{9cm}{head non-terminal probability\\ $p_{\bm{z}}(B, \curvearrowleft \mid A, \alpha)$ or $p_{\bm{z}}(C, \curvearrowright\mid A, \alpha)$:}}] Probability of the headedness direction and head-inheriting child\footnote{Child non-terminals that inherit the parent's head word.} conditioned on the parent non-terminal and head words.
\item[{\parbox[t]{9cm}{non-inheriting child probability\\ $p_{\bm{z}}(C \mid A, B, \alpha, \curvearrowleft)$ or $p_{\bm{z}}(B \mid A, C, \alpha, \curvearrowright)$:}}] Probability of the non-inheriting child conditioned on the headedness direction, and parent and head-inheriting child non-terminals. 
\end{description}

The score of root-seeking rule \bluecircled{1} is factorized as the product of the root to non-terminal score and word emission scores, as shown in \Cref{eq:rule1}.
\begin{equation}
\small
\label{eq:rule1}
g_{\theta}(S\rightarrow A[\alpha], \bm{z}) = \log p_{\bm{z}}(S\rightarrow A) + \log p_{\bm{z}}(A\rightarrow \alpha)
\end{equation}

The scores of branching rules \yellowcircled{2l} and \yellowcircled{2r} are factorized as the sum of a binary non-terminal score, a head non-terminal score, and a word emission score.
\Cref{eq:rule2} describes the factorization of the score of rule \yellowcircled{2l} and \yellowcircled{2r}:
\begin{equation}
\small
    \label{eq:rule2}
    \begin{aligned}
    g_{\theta}(&A[\alpha]\rightarrow B[\alpha]C[\beta], \bm{z})\\  
    &= \log p_{\bm{z}}(B, \curvearrowleft \mid A, \alpha) + \log p_{\bm{z}}(C \mid A, B, \alpha, \curvearrowleft)\\  
    &+ \log p_{\bm{z}}(C\rightarrow \beta)\\
    g_{\theta}(&A[\alpha]\rightarrow B[\beta]C[\alpha], \bm{z})\\  
    &= \log p_{\bm{z}}(C, \curvearrowright\mid A, \alpha) + \log p_{\bm{z}}(B\mid A, C, \alpha, \curvearrowright)\\  
    &+ \log p_{\bm{z}}(B\rightarrow \beta)
    \end{aligned}
\end{equation}

Since the head of preterminals is already specified upon generation of one of the ancestor non-terminals, the score of emission rule \redcircled{3} is 0. 

The component probabilities are all similarly parameterized, vectors corresponding to component non-terminals or terminals are fed through a multi-layer perceptron denoted $f(\cdot)$, and a dot product is taken with another vector corresponding to a component non-terminal or terminal.
Specifically, the root to non-terminal probability is
\begin{equation}
\begin{aligned}
    p_{\bm{z}}(S\rightarrow A) &= \exp \pi_{\bm{z}}(S\rightarrow A) / Z(\bm{z}),\\
    \pi_{\bm{z}}(S\rightarrow A) &= f^{(1)}([\bm{u}_S; \bm{z}])^T \bm{v}_A,
\end{aligned}
\end{equation}
where $;$ denotes concatenation and the word emission probability is
\begin{equation}
\begin{aligned}
    p_{\bm{z}}(A\rightarrow \alpha) &= \exp \pi_{\bm{z}}(A\rightarrow \alpha) / Z(A, \bm{z})\\
    \pi_{\bm{z}}(A\rightarrow \alpha) &= f^{(2)}([\bm{u}_A; \bm{z}])^T \bm{v}_{\alpha},
\end{aligned}
\end{equation}
with partition functions $Z(\bm{z})$ s.t. $\sum_{A\in \mathcal{N}} p_{\bm{z}}(S\rightarrow A) = 1$ and $Z(A, \bm{z})$ s.t. $\sum_{\alpha\in\Sigma}p_{\bm{z}}(A\rightarrow \alpha)=1$.

    The non-inheriting child probabilities for left- and right-headed dependencies are
\begin{equation}
\begin{aligned}
    p_{\bm{z}}(C \mid A, B, \alpha, \curvearrowleft) &= \frac{\exp \pi_{\bm{z}}(A[\alpha]\overset{\curvearrowleft}{\rightarrow} B[\alpha]C)}{Z(A, B, \alpha, \curvearrowleft, \bm{z})}\\
    p_{\bm{z}}(B \mid A, C, \alpha, \curvearrowright) &= \frac{\exp\pi_{\bm{z}}(A[\alpha]\overset{\curvearrowright}{\rightarrow} BC[\alpha])}{Z(A, C, \alpha, \curvearrowleft, \bm{z})}\\
    \pi_{\bm{z}}(A[\alpha]\overset{\curvearrowleft}{\rightarrow} B[\alpha]C) &= [\bm{w}_A^{\curvearrowleft};\bm{w}_{\alpha}^{\curvearrowleft}; \bm{z}]^T\bm{v}_{BC} \\
    \pi_{\bm{z}}(A[\alpha]\overset{\curvearrowright}{\rightarrow} BC[\alpha]) &= [\bm{w}_A^{\curvearrowright};\bm{w}_{\alpha}^{\curvearrowright}; \bm{z}]^T\bm{v}_{BC}
\end{aligned}
\end{equation}
where partition functions satisfy $\sum_{C}p_{\bm{z}}(C \mid A, B, \alpha, \curvearrowleft)$ and $\sum_{B}p_{\bm{z}}(B \mid A, C, \alpha, \curvearrowright) = 1$.

The respective head non-terminal scores are
\begin{equation}
\begin{aligned}
     p_{\bm{z}}(B, \curvearrowleft \mid A, \alpha) &= \frac{\exp \pi_{\bm{z}}(A[\alpha]\overset{\curvearrowleft}{\rightarrow}B[\alpha])}{Z(A, \alpha, \bm{z})}\\
    p_{\bm{z}}(C, \curvearrowright\mid A, \alpha) &=
    \frac{\exp \pi_{\bm{z}}(A[\alpha]\overset{\curvearrowright}{\rightarrow}B[\alpha])}{Z(A, \alpha, \bm{z})}\\
    \pi_{\bm{z}}(A[\alpha]\overset{\curvearrowleft}{\rightarrow}B[\alpha]) &= f^{(3)}([\bm{u}_A; \bm{u}_{\alpha};\bm{z}])^T\bm{v}_{B\curvearrowleft}  \\
    \pi_{\bm{z}}(A[\alpha]\overset{\curvearrowright}{\rightarrow}B[\alpha]) &= f^{(3)}([\bm{u}_A; \bm{u}_{\alpha};\bm{z}])^T\bm{v}_{B\curvearrowright}.
\end{aligned}
\end{equation}
where the partition function satisfies $\sum_{B}p_{\bm{z}}(B, \curvearrowleft \mid A, \alpha) + \sum_{C}p_{\bm{z}}(C, \curvearrowright \mid A, \alpha) = 1$.

Here vectors $\bm{u}, \bm{v}, \bm{w} \in \mathbb{R}^d$ represent the embeddings of non-terminals, preterminals and words. $f^{(i)}, i = 1, 2, 3$ are multi-layer perceptrons with different set of parameters, where we use residual connections\footnote{$f(x) = \sigma(W_2(\sigma(W_1 x + b_1)) + b ) + x$} \cite{he2016deep} between layers to facilitate training of deeper models. 
\subsection{Compound Grammar}
\label{sec:compound_grammar}
Among various existing grammar induction models, the compound PCFG model of \citep{kim2019compound} both shows highly competitive results and follows a PCFG-based formalism similar to ours, and thus we build upon this method. The \textit{compound} in compound PCFG refers to the fact that it uses a compound probability distribution \cite{robbins1951asymptotically} in modeling and estimation of its parameters. A compound probability distribution enables continuous variants of grammars, allowing the probabilities of the grammar to change based on the unique characteristics of the sentence. 
In general, compound variables can be devised in any way that may inform the specification of the rule probabilities (e.g. a structured variable to provide frame semantics or the social context in which the sentence is situated).
In this way, compound grammar increases the capacity of the original PCFG.

In this paper, we use a latent compound variable $\bm{z}$ which is sampled from a standard spherical Gaussian distribution.  
\begin{equation}
    \bm{z} \sim \mathcal{N}(\mathbf{0}, \mathbf{I})
\end{equation}

We denote the probability of latent variable $\bm{z}$ as $p_{\mathcal{N}(\mathbf{0}, \mathbf{I})}(\bm{z})$. By marginalizing out the compound variable, we get the log likelihood of a sentence:
\begin{equation}
    \log p(\bm{x}) = \log \int_{\bm{z}} p_{\bm{z}}(\bm{x}) p_{\mathcal{N}(\mathbf{0}, \mathbf{I})}(\bm{z}) \dif \bm{z}
\end{equation}

\begin{algorithm}[!t]
\begin{algorithmic}
\Require $\mathcal{N}, \mathcal{T}, P_1, P_2$
\Function{recursive}{$N, \alpha, \bm{z}$}
    \State $N_1, N_2, d, \beta \sim P_2(N_1, N_2, d, \beta\mid N, \alpha, \bm{z})$
    \If{$d=\curvearrowright$}
        \State $\alpha, \beta \leftarrow \beta, \alpha$
    \EndIf
    \If{$N_1\in\mathcal{N}$}
        \State $\mathcal{S}_l \leftarrow $ {\sc recursive}($N_1, \alpha, \bm{z}$)
    \Else
        \State $\mathcal{S}_l \leftarrow [\alpha]$
    \EndIf
    \If{$N_2\in\mathcal{N}$}
        \State $\mathcal{S}_r \leftarrow $ {\sc recursive}($N_2, \beta, \bm{z}$)
    \Else
        \State $\mathcal{S}_r \leftarrow [\beta]$
    \EndIf
    \State \textbf{return} {\sc concatenate}($S_l, S_r$)
\EndFunction
\State $\bm{z} \sim \mathcal{N}({\mathbf{0}, \mathbf{I}})$
\State $N, \alpha \sim P_1(N, \alpha\mid S, \bm{z})$
\State \textbf{return} {\sc recursive}($N, \alpha, \bm{z}$)
\end{algorithmic}
\label{alg:gen}
\caption{Generative Procedure of Neural L-PCFGs: Sentences Are Generated from Start Symbol $S$ and Compound Variable $\bm{z}$ Recursively.}
\end{algorithm}

\Cref{alg:gen} shows the procedure to generate a sentence recursively from a random compound variable and a distribution over the production rules in a pre-order traversal manner, where $P_1$ and $P_2$ are defined using $g_\theta$ from Eqs (\ref{eq:rule1}) and (\ref{eq:rule2}), respectively:
\begin{equation}
\small
    \begin{aligned}
        P_1(N, \alpha \mid S, \bm{z}) &= \exp(g_{\theta}(S\rightarrow A[\alpha], \bm{z})) \\
        P_2(N_1, N_2, \curvearrowleft, \beta&\mid N, \alpha, \bm{z})\\ &= \exp(g_{\theta}(A[\alpha]\rightarrow B[\alpha]C[\beta], \bm{z})) \\
        P_2(N_1, N_2, \curvearrowright, \beta&\mid N, \alpha, \bm{z})\\
        &= \exp(g_{\theta}(A[\alpha]\rightarrow B[\beta]C[\alpha], \bm{z}))
    \end{aligned}
\end{equation}
\section{Training and Inference}


\subsection{Training}
\label{sec:training}
It is intractable to obtain either the exact log likelihood by integration over $z$, and estimation by Monte-Carlo sampling would be hopelessly inefficient. However, we can optimize the evidence lower bound (ELBo):
\begin{equation}
    \begin{aligned}
    \mathcal{L}(\bm{x}) & = \mathbb{E}_{q_{\phi}(\bm{z}\mid \bm{x})}\log  p_{\bm{z}}(\bm{x})\\
    &\qquad- \text{KL}[q_{\phi}(\bm{z}\mid \bm{x}) \| p_{\mathcal{N}(\mathbf{0}, \mathbf{I})}(\bm{z})] \\
    & \leq \mathbb{E}_{p_{\mathcal{N}(\mathbf{0}, \mathbf{I})}(\bm{z})} p_{\bm{z}}(\bm{x})
    \end{aligned}
\end{equation}
where $q_{\phi}(\bm{z}\mid \bm{x})$ is the proposal probability parameterized by an inference network, similar to those used in variantial autoencoders \cite{kingma2013autoencoding}. The ELBo can be estimated by Monte-Carlo sampling:
\begin{equation}
 \begin{aligned}
    \mathcal{L}(x) = &\frac{1}{L} \sum_{i=1}^L \log p_{\bm{z}_i}(\bm{x}) \\
    &- \text{KL}[q_{\phi}(\bm{z}\mid \bm{x}) \| p_{\mathcal{N}(\mathbf{0}, \mathbf{I})}(\bm{z})]
\end{aligned}
\end{equation}
where $\{\bm{z}_i\}_{i=1}^L$ are sampled from $q_{\phi}(\bm{z}\mid \bm{x})$. We model the proposal probability as an orthogonal Gaussian distribution:
\begin{equation}
    q_{\phi}(\bm{z}\mid\bm{x}) = p_{\mathcal{N}(\bm{\mu}, \text{diag}(\bm{\sigma}))}(\bm{z})
\end{equation}
where ($\bm{\mu}$, $\bm{\sigma}$) are output by the inference network
\begin{equation}
    \bm{\mu} = f_{\mu}(\bm{x}), \bm{\sigma} = f_{\sigma}(\bm{x})
\end{equation}
Both $f_{\mu}$ and $f_{\sigma}$ are parameterized as LSTMs \cite{hochreiter1997long}.
Note that the inference network could be optimized by the reparameterization trick \cite{kingma2013autoencoding}:
\begin{equation}
    \hat{\bm{z}} \sim \mathcal{N}(\mathbf{0}, \mathbf{I}), \bm{z} = \bm{\mu} + \bm{\sigma} \odot \hat{\bm{z}}
\end{equation}
where $\odot$ denotes Hadamard operation.
The KL divergence between $q_{\phi}(\bm{z}\mid\bm{x})$ and $\mathcal{N}(\mathbf{0}, \mathbf{I})$ is
\begin{equation}
\begin{aligned}
    \text{KL}[&q_{\phi}(\bm{z}\mid\bm{x}) \| p_{\mathcal{N}(\mathbf{0}, \mathbf{I})}(\bm{z})]\\ 
    &= -\frac{1}{2}(\sum_{i=1}^n(\log \sigma_i - \sigma_i + 1) - \|\bm{\mu}\|_2^2).
\end{aligned}
\end{equation}
\begin{figure}[!t]
    \centering
    \resizebox{\linewidth}{!}{
    \includegraphics[width=\linewidth]{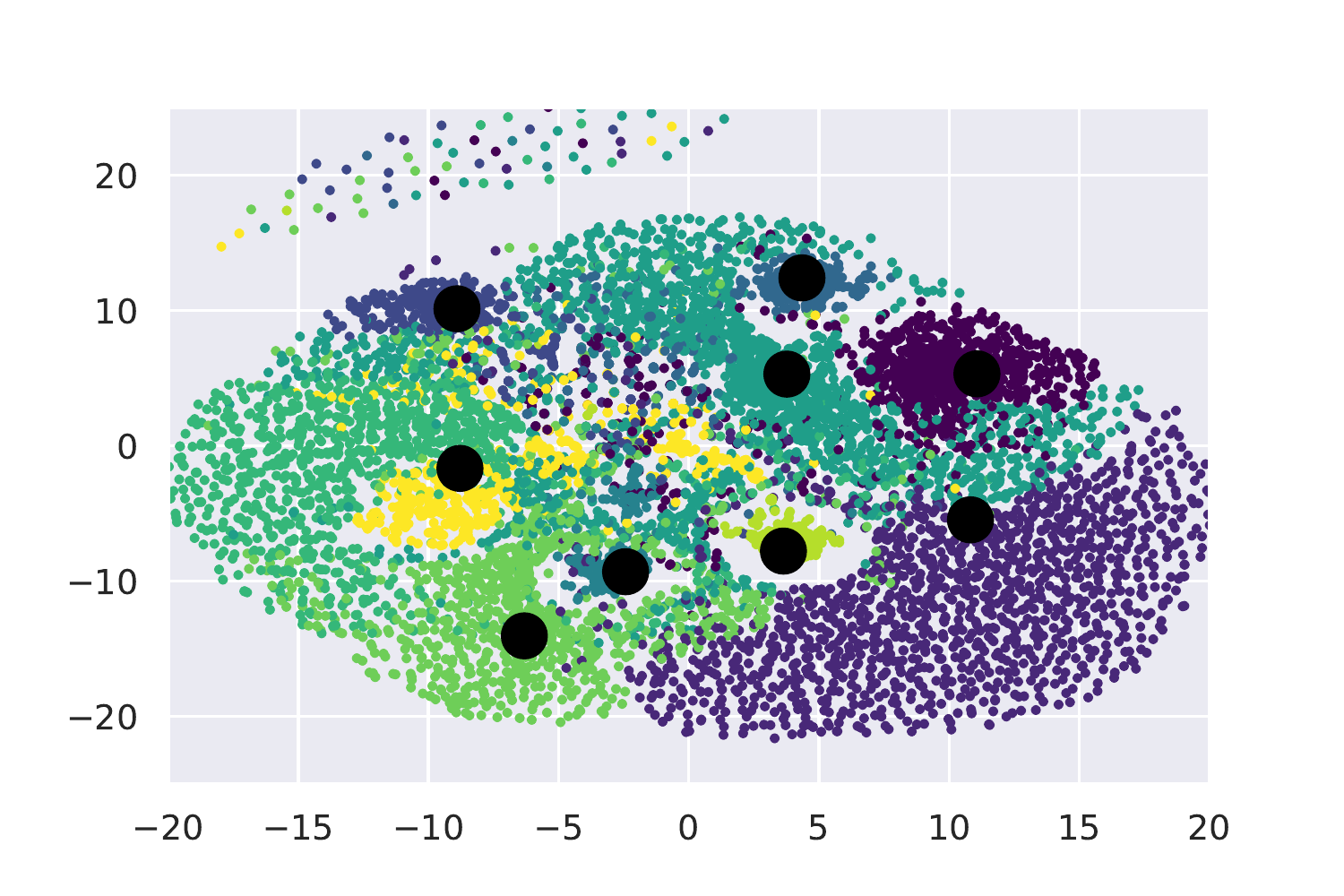}}
    \caption{The clustering Result of GloVe Embeddings. Different colors represent cluster class of each word, and larger black points represent the initial embeddings of preterminals, i.e cluster centroids. The two-dimension visualization is obtained by TSNE \cite{maaten2008visualizing}.}
    \label{fig:cluster}
\end{figure}
\paragraph{Initialization} We initialize word embeddings using GloVe embeddings \cite{pennington2014GloVe}. We further cluster word embeddings with K-Means \cite{macqueen1967some}, as shown in \Cref{fig:cluster} and use the centroids of the clusters to initialize the embeddings of preterminals. The K-Means algorithm is initialized using the K-Means++ method and trained until convergence. 
The intuition therein is that this gives the model a rough idea of syntactic categories before starting grammar induction.
We also consider the variant without pretrained word embeddings, where we initialize word embeddings and preterminals both by drawing from $\mathcal{N}(\bm{0}, \bm{I})$. Other parameters are initialized by Xavier normal initialization \cite{glorot2010understanding}.

\paragraph{Curriculum Learning} We also apply curriculum learning \cite{bengio2009curriculum,Spitkovsky2010} to learn the grammar gradually. Starting at half of the maximum length in the training set, we raise the length limit by $\alpha\%$ each epoch. 

\subsection{Inference}
\label{sec:inference}
We are interested in the induced parse tree for each sentence in the task of unsupervised parsing, i.e. the most probable tree $\hat{t}$ 
\begin{equation}
    \begin{aligned}
    \hat{t} &= \arg\max_t p(t\mid \bm{x})\\
    &= \arg\max_{t\in\mathcal{T}_x} \int_{\bm{z}} p_{\bm{z}}(t) p(\bm{z}\mid\bm{x}) \dif \bm{z}
    \end{aligned}
\end{equation}
where $p(\bm{z}\mid\bm{x})$ is the posterior over compound variables.
However, it is intractable to get the most probable tree. Hence we use the mean $\bm{\mu} = f_{\mu}(\bm{x})$ predicted by the inference network and replace $p(\bm{z}\mid\bm{x})$ with a Dirac delta distribution $\delta(\bm{z} - \bm{\mu})$ in place of the real distribution to approximate the integral\footnote{Note that it is also possible to use other methods for approximation. For example, we can use $q_{\phi}(\bm{z}\mid\bm{x})$ in place of posterior distribution. However, using it still results in high prediction variance of the max function. We did not observe a significant improvement with other methods.} 
\begin{equation}
    \begin{aligned}
    \hat{t} &\approx \arg\max_{t\in\mathcal{T}_x} \int_{\bm{z}} p_{\bm{z}}(t) \delta(\bm{z} - \bm{\mu}) \dif \bm{z}\\
    &= \arg\max_{t\in\mathcal{T}_x} p_{\bm{\mu}}(t)
    \end{aligned}
\end{equation}

The most probable tree can be obtained via CYK algorithm.

\section{Experiments}
\subsection{Data Setup}
All models are evaluated using the Penn Treebank \cite{marcus1993building} as the test corpus, following the splits and preprocessing methods, including removing punctuation, provided by \citet{kim2019compound}. To convert the original phrase bracket and label annotations to dependency annotations, we use Stanford typed dependency representations \cite{de2008stanford}. 

We employ three standard metrics to measure the performance of models on the validation and test sets: directed and undirected attachment score (DAS and UAS) for dependency parsing, and unlabeled constituent F1 score for constituency parsing.

We tune hyperparameters of the model to minimize perplexity on the validation set.
We choose perplexity because it requires only plain text and not annotated parse trees.
Specifically, we tuned the architecture of $f^{(i)}, i=1, 2, 3$ in the space of multilayer perceptrons, with the dimension of each layer being $n+d$, with residual connections and different non-linear activation functions.
\Cref{tab:hyperparameters} shows the final hyper-parameters of our model. 
Due to memory constraints on a single graphic card, we set the number of non-terminals and preterminals to 10 and 20, respectively. Later we will show that the compound PCFG's performance is benefited by a larger grammar, it is therefore possible the same is true for our neural L-PCFG.
\Cref{sec:conclusion} includes a more detailed discussion of space complexity.

\begin{table}[!t]
    \centering
    \begin{tabular}{c|c}
    \toprule
    Hyperparameter     &  Value\\ \midrule
    $|\mathcal{N}|$, $|\mathcal{P}|$      & 10, 20\\
    $n$ & 60 \\
    $d$ & 300 \\
    $\alpha$ & 10 \\
    \#layers of $f^{(1)}$, $f^{(2)}$, $f^{(3)}$  & 6, 6, 4\\
    non-linear activation & \texttt{relu}\\
    \bottomrule
    \end{tabular}
    \caption{Hyper-parameters and values}
    \vspace{-2mm}
    \label{tab:hyperparameters}
\end{table}

\newcommand{\Be}[1]{\textbf{#1}}
\begin{table*}[!t]
\centering
\resizebox{\linewidth}{!}{
\begin{tabular}{@{}lcc|rrrrrr@{}}
 \multicolumn{1}{c}{}& \multicolumn{1}{c}{}& \multicolumn{1}{c}{} & \multicolumn{2}{c}{DAS} & \multicolumn{2}{c}{UAS} & \multicolumn{2}{c}{F1}    \\ \cmidrule(l){4-5} \cmidrule(l){6-7} \cmidrule(l){8-9} 
\multicolumn{1}{c}{}& \multicolumn{1}{c}{Gold Tags} & \multicolumn{1}{c}{Word Embedding} & \multicolumn{1}{c}{Dev}        & \multicolumn{1}{c}{Test}       & \multicolumn{1}{c}{Dev}        & \multicolumn{1}{c}{Test}  & \multicolumn{1}{c}{Dev} & \multicolumn{1}{c}{Test}    \\ 
\toprule
Compound PCFG$^{**}$  & \xmark  & $\mathcal{N}(\mathbf{0}, \mathbf{I})$ &    21.2 \phantom{(0.0)}       &  23.5 \phantom{(0.0)}        &  38.9 \phantom{(0.0)}      &    40.8 \phantom{(0.0)}  &  \multicolumn{1}{c}{-} &  \Be{55.2}\phantom{0} \phantom{(0.00)}\\
\midrule\midrule
Compound PCFG & \xmark  & $\mathcal{N}(\mathbf{0}, \mathbf{I})$ &       15.6 (3.9)     &         17.8 (4.2)  &        27.7 (4.1)  &     30.2 (5.3)  & 45.63 (1.71) & 47.79 (2.32)  \\
Compound PCFG & \xmark & GloVe &    16.4 (2.4)       &  18.6 (3.7)     &      28.7 (3.5)   &    31.6 (4.5)    & 45.52 (2.14) &  48.20 (2.53) \\
DMV & \xmark & - & 24.7 (1.5) & 27.2 (1.9) & 43.2 (1.9) & 44.3 (2.2)  & \multicolumn{1}{c}{-} & \multicolumn{1}{c}{-} \\
DMV & \cmark & - & 28.5 (1.9)  &    29.9 (2.5)       &    45.5 (2.8)        &  47.3 (2.7)    &  \multicolumn{1}{c}{-} &  \multicolumn{1}{c}{-}    \\
 \midrule
Neural L-PCFGs & \xmark & $\mathcal{N}(\mathbf{0}, \mathbf{I})$  &     37.5 (2.7)       &     39.7 (3.1)      &     50.6 (3.1)      &  53.3 (4.2)  & \Be{52.90 (3.72)} &  \Be{55.31 (4.03)}  \\
Neural L-PCFGs & \xmark & GloVe &  \Be{38.2 (2.1)}     &     \Be{40.5 (2.9)}   &     \Be{54.4 (3.6)}      &    \Be{55.9 (3.8)}    & 45.67 (0.95) &  47.23 (2.06)  \\
Neural L-PCFGs &\cmark & $\mathcal{N}(\mathbf{0}, \mathbf{I})$ &       35.4 (0.5)     &        39.2 (1.1)    &        50.0 (1.3)    &    53.8 (1.7)   & 51.16 (5.11) & 54.49 (6.32)  \\ 
\bottomrule
\end{tabular}}
\caption{Dependency and constituency parsing results. 
\textit{DAS}/\textit{UAS} stand for directed/undirected accuracy.  For the compound PCFG we use heuristic head rules to obtain dependencies (\S\ref{sec:baselines}). Figures in the parenthesis show the standard deviation calculated from five runs with different random seeds. $^{**}$indicates a large (30 NT, 60 PT) compound PCFG from \citet{kim2019compound} -- we could not use this size in our experiments due to memory constraints. Results are not directly comparable with the other rows due to model size, but we report them for completeness. Best average performances are indicated in \textbf{bold}.
}
\vspace{-2mm}
\label{tab:result}
\end{table*}

\subsection{Baselines}
\label{sec:baselines}
We compare our neural L-PCFGs with the following baselines:

\paragraph{Compound PCFG} The compound PCFG \citep{kim2019compound} is an unsupervised constituency parsing model which is a PCFG model with neural scoring. The main difference between this model and neural L-PCFG is the modeling of headedness and the dependency between head word and generated non-terminals or preterminals. We apply the same hyperparameters and techniques, including number of non-terminals and preterminals, initialization, curriculum learning and variational
training to compound PCFGs for a fair comparison.
Because compound PCFGs have no notion of dependencies, 
we extract dependencies from the compound PCFG with three kinds of heuristic head rules: left-headed, right-headed and large-headed.
Left-/right-headed mean always choosing the root of the left/right child constituent as the root of the parent constituent, whereas large-headedness is generated by a heuristic rule which chooses the root of larger child constituent as the root of the parent constituent.
Among these, we choose the method that obtains the best parsing accuracy on the dev set (making these results an oracle with access to \emph{more} information than our proposed method).

\paragraph{Dependency Model with Valence (DMV)} The DMV \cite{klein2004corpus} is a model for unsupervised dependency parsing, where \textit{valence} stands for the number of arguments controlled by a head word. The choices to attach words as children are conditioned on the head words and valences. 
As shown in \citep{smith2006novel}, the DMV model can be expressed as a head-driven context-free grammar with a set of generation rules and scores, where the non-terminals represent the valence of head words.
For example, ``L[{\sc chasing}] $\rightarrow$ $\text{L}_0$[{\sc is}] R[{\sc chasing}]'' denotes that left-hand constituent with full left valence produces a word and a constituent with full right valence.
Therefore, it could be seen as a special case of lexicalized PCFG where the generation rules provide inductive biases for dependency parsing but are also restricted -- for example, a void-valence constituent cannot produce a full-valence constituent with the same head.

Note that DMV uses far fewer parameters than the PCFG based models, $\mathcal{O}(|\mathcal{P}|^2)$. The neural L-PCFG uses a similar number of parameters as we do, $\mathcal{O}(n(|\mathcal{P}| + |\mathcal{N}|) + n^2)$.

We compare models under two settings: (1) with gold tag information and (2) without it, denoted by \cmark and \xmark, respectively in \Cref{tab:result}. To use gold tag information in training the neural L-PCFG, we assign the 19 most frequent tags as categories and combine the rest into a 20th ``other'' category. These categories are used as supervision for the preterminals. In this setting, instead of optimizing the log probability of the sentence, we optimize the log joint probability of the sentence and the tags.

\subsection{Quantitative Results}
\label{sec:effectiveness}

First, in this section, we present and discuss quantitative results, as shown in \Cref{tab:result}.

\subsubsection{Main Results}


First comparing neural L-PCFGs with compound PCFGs, we can see that L-PCFGs perform slightly better on phrase structure prediction and achieve much better dependency accuracy. This shows that (1) lexical dependencies contribute somewhat to the learning of phrase structure, and (2) the head rules learned by neural L-PCFGs are significantly more accurate than the heuristics that we applied to standard compound PCFGs. We also find that GloVe embeddings can help (unsupervised) dependency parsing, but do not benefit constituency parsing.


Next, we can compare the dependency induction accuracy of the neural L-PCFGs with the DMV.
The results indicate that neural L-PCFGs without gold tags achieve even better accuracy than DMV with gold tags on both directed accuracy and undirected accuracy.
As discussed before, DMV can be seen as a special case of L-PCFG where the attachment of children is conditioned on the valence of the parent tag, while in L-PCFG the generated head directions are conditioned on the parent non-terminal and the head word, which is more general.
Comparatively positive results show that conditioning on generation rules not only is more general but also yields a better prediction of attachment. 

\Cref{tab:recall_by_label} shows label-level recall, i.e. unlabeled recall of constituents annotated by each non-terminal. We observe that the neural L-PCFG outperforms all baselines on these frequent constituent categories. 

\begin{table}[]
\small
    \centering
    \resizebox{\linewidth}{!}{
    \begin{tabular}{l c c c c}
    \toprule
    & \multirow{2}{*}{PRPN}& \multirow{2}{*}{ON} & Compound  & Neural\\
    &  &  & PCFG  & L-PCFG\\
    \midrule 
    SBAR & 50.0\% & 51.2\% & 42.36\%  & \Be{53.60}\%\\
    NP & 59.2\% & 64.5\% & 59.25\%  & \Be{67.38}\% \\
    VP & 46.7\% & 41.0\% & 39.50\%  & \Be{48.58}\%\\
    PP & 57.2\% & 54.4\% & 62.66\%  & \Be{65.25}\%\\
    ADJP & 44.3\% & 38.1\% & 49.16\%  & \Be{49.83}\%\\
    ADVP & 32.8\% & 31.6\% & 50.58\%  & \Be{58.86}\% \\
         \bottomrule
    \end{tabular}}

    \caption{Fraction of ground truth constituents that were predicted as a constituent by the models broken down by label (i.e. label recall). Results of PRPN and ON are from \cite{kim2019compound}.}
    \label{tab:recall_by_label}
\end{table}

\subsubsection{Impact of Factorization}
\label{sec:impact_of_factorization}
\label{sec:ablation_study}
\Cref{tab:ablation} compares the effects of three alternate factorizations of $g_{\theta}(A[\alpha]\rightarrow B[\alpha]C[\beta], \bm{z})$:
\vspace{-2mm}
\begin{center}
    \begin{tabular}{@{}l@{\hspace{2pt}}l@{\hspace{-1em}}l}
    \multicolumn{3}{@{}l}{$g_{\theta}(A[\alpha]\rightarrow B[\alpha]C[\beta], \bm{z}) = p_{\bm{z}}(C\rightarrow \beta)\hspace{3pt}+$}\\
\textbf{F I}:   &$  \log p_{\bm{z}}(B, C \mid A)$ & $+  \log p_{\bm{z}}(\curvearrowleft\mid A\rightarrow BC)$ \label{eq:fac1} \\ 
\textbf{F II}:  &$  \log p_{\bm{z}}(B, C, \curvearrowleft\mid A, \alpha)$  \label{eq:fac2} & \\
\textbf{F III}: &$  \log p_{\bm{z}}(B, \curvearrowleft \mid A, \alpha) $ & $+\log p_{\bm{z}}(C\mid A, B, \alpha)\notag$\label{eq:fac3} 
        \end{tabular}
\end{center}

Factorization I assumes that the child non-terminals do not depend on the head lexical item, which influences the parsing result significantly. Although Factorization II is as general as our proposed method, it uses separate representations for different directions, $\bm{v}_{BC\curvearrowleft}$ and $\bm{v}_{BC\curvearrowright}$. Factorization III assumes the independence between direction and dependent non-terminals. These results indicate that our factorization strikes a good balance between modeling lexical dependencies and directionality, and avoiding over-parameterization of the model that may lead to sparsity and difficulties in learning.

\begin{table}[!t]
\centering
\resizebox{\linewidth}{!}{
\begin{tabular}{@{}lrrr@{}}
\toprule
 \multicolumn{1}{c}{}&  DAS & UAS & F1    \\  \midrule
Neural L-PCFG &      35.5             &     51.4           & 44.5 \\
\qquad w/ \textbf{xavier init} &     27.2               &     47.6       & 43.6  \\
\qquad w/ \textbf{Factorization I} & 16.4  & 33.3 & 25.7  \\
\qquad w/ \textbf{Factorization II} & 22.3 & 42.7 &  39.6 \\
\qquad w/ \textbf{Factorization III} & 25.9 & 46.9 & 34.7\\\bottomrule
\end{tabular}}
\caption{An ablation of dependency and constituency parsing results on the validation set with different settings of neural L-PCFG. All models are trained with GloVe word embeddings and without gold tags. ``w/ \textbf{xavier init}'' means that preterminals are not initialized by clustering centroids by xavier normal distribution. ``w/ Factorization N'' represents different factorization methods (\S \ref{sec:ablation_study}).}
\label{tab:ablation}
\vspace{-3mm}
\end{table}

\subsection{Qualitative Analysis}
\label{sec:qualitative_analysis}
\begin{figure}[!t]
    \centering
    \begin{subfigure}{\linewidth}
    \includegraphics[width=\linewidth]{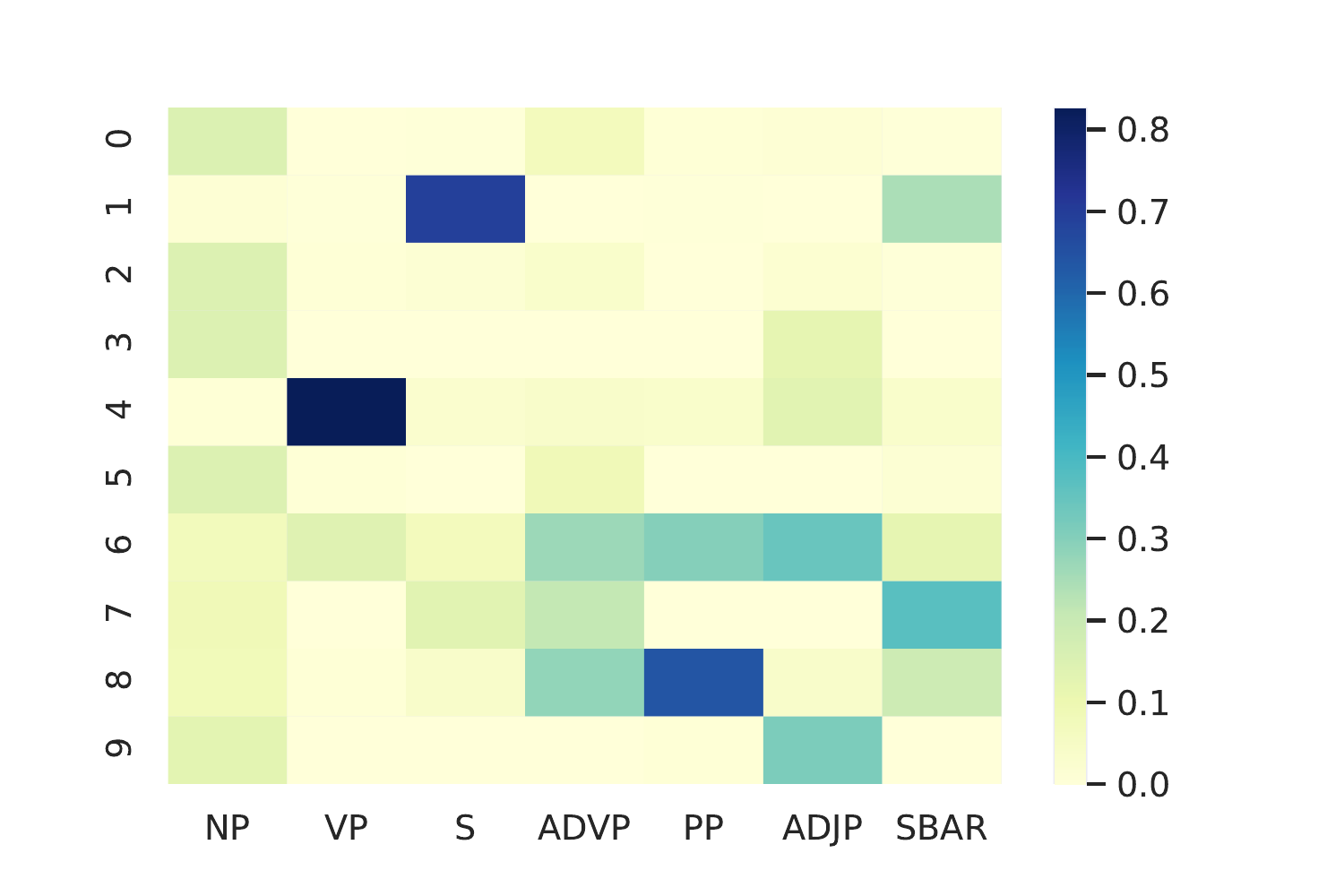}
    \end{subfigure}
    \vspace{-2mm}
    \begin{subfigure}{\linewidth}
    \includegraphics[width=\linewidth]{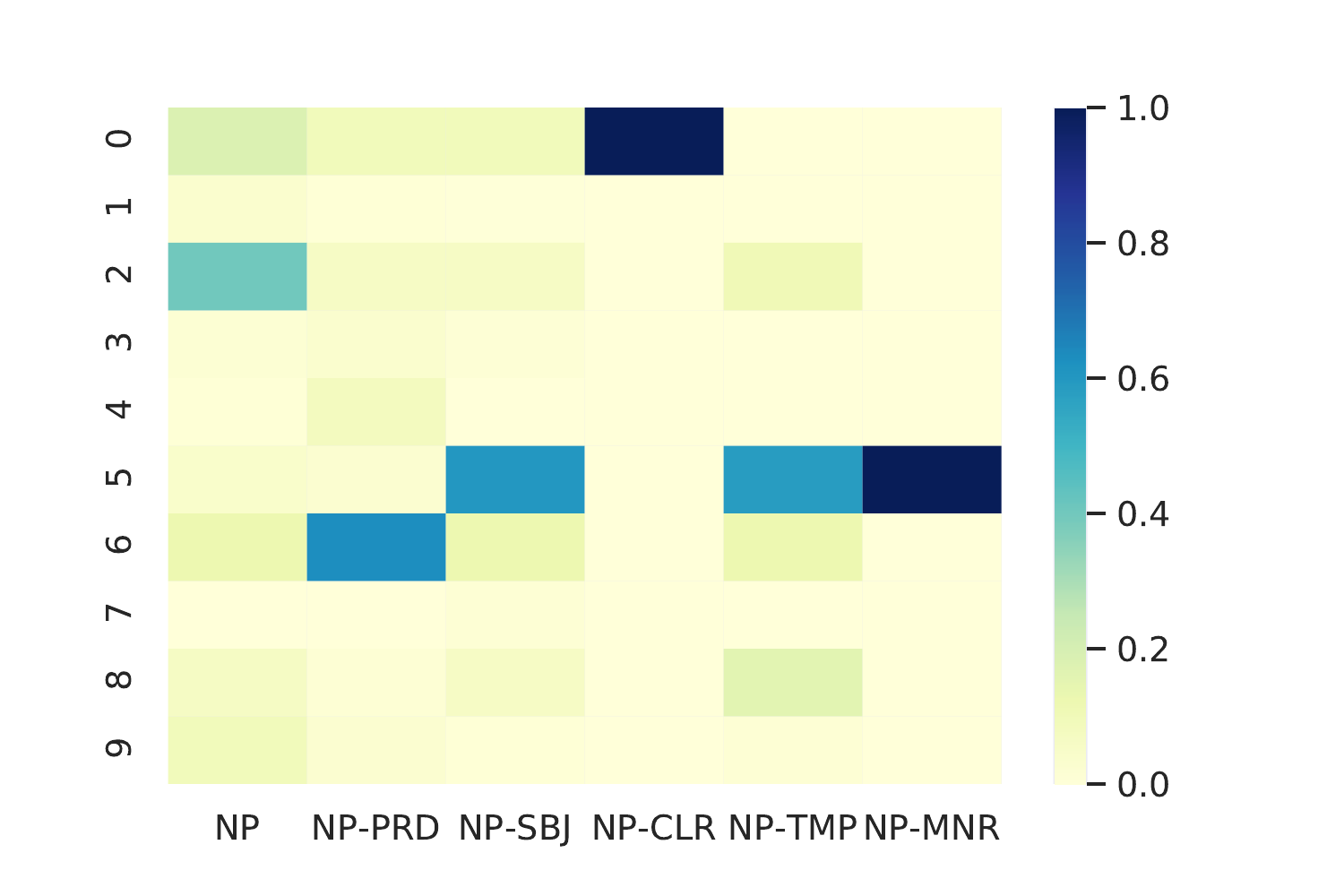}
    \end{subfigure}
    \caption{Alignment between all induced non-terminals (x-axis) and gold non-terminals annotated in the PTB (y-axis). In the upper figure, we show the seven most frequent gold non-terminals, and list them by frequency from left to right. For each gold non-terminal, we show the proportion of each induced non-terminal. In the lower map, we breakdown the results of the noun phrase (NP) into subcategories. Darker color indicates higher proportion, and vice versa.}
    \label{fig:alignment_map}
\end{figure}

We analyze our best model without gold tags in detail. \Cref{fig:alignment_map} visualizes the alignment between our induced non-terminals and gold constituent labels on the overlapping constituents of induced trees and the ground-truth. For each constituent label, we show the frequency of it annotating the same span of each non-terminal. We observe from the first map that a clear alignment between certain linguistic labels and induced non-terminals, e.g. VP and NT-4, S and NT-2, PP and NT-8. But for other non-terminals, there's no clear alignment with induced classes. One hypothesis for this diffusion is due to the diversity of the syntactic roles of these constituents. To investigate this, we zoom in on noun phrases in the second map, and observe that NP-SBJ, NP-TMP and NP-MNR are combined into a single non-terminal NT-5 in the induced grammar, and that NP, NP-PRD and NP-CLR corresponds to NT-2, NT-6 and NT-0, respectively. 

We also include an example set of parses for comparing the DMV and neural L-PCFG in \Cref{tab:case_study_1}. Note that DMV uses ``to'' as the head of ``know'', the neural L-PCFG correctly inverts this relationship to produce a parse that is better aligned with the gold tree. One of the possible reasons that the DMV tends to use ``to'' as the head is that DMV has to carry the information that the verb is in the infinitive form, which will be lost if it uses ``know'' as the head. In our model, however, such information is contained in the types of non-terminals. In this way, our model uses the open class word ``know'' as the root. Note that we also illustrate a similar failure case in this example. Neural L-PCFG uses ``if'' as the head of the if-clause, which is probably due to the independency between the root of the if-clause and ``know''.

A common mistake made by the neural L-PCFG is treating auxiliary verbs like adjectives that combine with the subject instead of modifying verb phrases. For example, the neural L-PCFG parses \textit{``...the exchange will look at the performance...''} as ``((the exchange) will) (look (at (the performance)))'', while the compound PCFG produces the correct parse ``((the exchange) (will (look (at (the performance))))''. A possible reason for this mistake is English verb phrases are commonly left-headed which makes attaching an auxiliary verb less probable as the left child of a verb phrase. This type of error may stem from the model's inability to assess the semantic function of auxiliary verbs \cite{Bisk2015}.

\begin{table*}[!t]
\centering
\resizebox{\textwidth}{!}{
\begin{tabular}{@{}p{25pt}c@{}}
\toprule
\begin{tabular}{@{}l@{}}DMV\end{tabular}&
\begin{tabular}{@{}l@{}}
\begin{dependency}[theme = simple]
   \begin{deptext}[column sep=1em]
      It \& 's \& hard \& to \& know \& if \& people \& are \& responding \& truthfully \\
   \end{deptext}
   \deproot[edge height=0.85cm]{2}{ROOT}
   \depedge{2}{1}{}
   \depedge{2}{3}{}
   \depedge{2}{4}{}
   \depedge{4}{5}{}
   \depedge{5}{6}{}
   \depedge{8}{7}{}
   \depedge{6}{8}{}
   \depedge{8}{9}{}
   \depedge{9}{10}{}
\end{dependency}
\end{tabular}
\\\midrule
\begin{tabular}{@{}l@{}}Neural L-PCFG\end{tabular}&
\begin{tabular}{@{}l@{}}
\small

\begin{tikzpicture}[level distance=20pt]
\Tree [.NT-8{\sc ['s]} [.T-10{\sc [It]} It ] [.NT-4{\sc ['s]} [.T-20{\sc ['s]} 's ] [.NT-7{\sc [hard]} [.T-6{\sc [hard]} hard ]  [.NT-10{\sc [know]} [.T-15{\sc [to]} to ] [.NT-5{\sc [know]} [.T-19{\sc [know]} know ] [.NT-1{\sc [if]} [.T-15{\sc [if]} if ] [.NT-9{\sc [responding]} [.T-10{\sc [people]} people ] [.NT-4{\sc [responding]} [.T-20{\sc [are]} are ] [.NT-8{\sc [responding]} [.T-11{\sc [responding]} responding ]  [.T-12{\sc [truthfully]} truthfully ] ] ] ] ] ] ] ] ] ]
\end{tikzpicture}
\end{tabular}
\\\midrule
\begin{tabular}{@{}p{4em}@{}}Dependencies from Neural L-PCFG\\ \end{tabular}&
\begin{tabular}{@{}l@{}}
\begin{dependency}[theme = simple]
   \begin{deptext}[column sep=1em]
      It \& 's \& hard \& to \& know \& if \& people \& are \& responding \& truthfully \\
   \end{deptext}
   \deproot[edge height=0.85cm]{2}{ROOT}
   \depedge{2}{1}{}
   \depedge{2}{3}{}
   \depedge{3}{5}{}
   \depedge{5}{4}{}
   \depedge{5}{6}{}
   \depedge{6}{9}{}
   \depedge{9}{7}{}
   \depedge{9}{8}{}
   \depedge{9}{10}{}
\end{dependency}
\end{tabular}
\\\midrule
\begin{tabular}{@{}l@{}}Gold Tree \end{tabular}&
\begin{tabular}{@{}l@{}}
\begin{dependency}[theme = simple]
   \begin{deptext}[column sep=1em]
      It \& 's \& hard \& to \& know \& if \& people \& are \& responding \& truthfully \\
   \end{deptext}
   \deproot[edge height=0.85cm]{2}{ROOT}
   \depedge{2}{1}{}
   \depedge{2}{3}{}
   \depedge{3}{5}{}
   \depedge{5}{4}{}
   \depedge{5}{9}{}
   \depedge{9}{6}{}
   \depedge{9}{7}{}
   \depedge{9}{8}{}
   \depedge{9}{10}{}
\end{dependency}
\end{tabular}\\\bottomrule
\end{tabular}
}
\caption{Comparison between Neural L-PCFG and DMV on a case from PTB training set. 
}
\label{tab:case_study_1}
\end{table*}


\section{Related Work}
\paragraph{Dependency vs Constituency Induction}
The decision to focus on modeling dependencies and constituencies has largely split the grammar induction community into two camps.  The most popular approach has been focused on dependency formalisms \cite{klein2004corpus,Spitkovsky2010,Spitkovsky2011a,Spitkovsky2013,marecek-straka-2013-stop, Jiang2016,Tran2018}, while a second community has focused on inducing constituencies \cite{lane2001incremental,Ponvert2011,golland-etal-2012-feature,jin-etal-2018-unsupervised}. Induced constituencies can in the case of CCG \cite{Bisk2012a,Bisk2013} produce dependencies, but unlike our proposal, existing approaches do not jointly model both representations. Context-free grammars (CFG) have been used for decades to represent, analyze and model the phrase structure of language \citep{chomsky1956syntactic,pullum1982natural,lari1990estimation,klein2002generative,bod2006all}. 

Similarly, the compound PCFG \citet{kim2019compound}, which we extend, falls into this camp of models that induce only phrase-structure grammar. However, in this paper we demonstrate a novel lexicaly informed neural parameterization which extends their model to induce a \emph{unified} phrase-structure and dependency-structure grammar.

\paragraph{Unifying Phrase Structure and Dependency Grammar}
Head-driven phrase structure grammar \citep{sag1987information} and lexicalized tree adjoining grammar \citep{schabes1988parsing} are 
approaches to representing dependencies directly in phrase structure.  

The notion that abstract syntactic structure should provide scaffolding for dependencies, and that lexical dependencies should provide a semantic guide for syntax, was most famously explored in \citet{collins2003head} through the introduction of an L-PCFG. In addition, \citet{carroll1998valence} explored the problem of head induction in L-PCFG; \citet{charniak2005coarse} improves L-PCFGs with coarse-to-fine parsing and reranking. Recently \cite{green2012hybrid,ren2013combine,yoshikawa2019ccg} explored various methods to jointly infer phrase structure and dependencies. 

\citet{klein2004corpus} show that a combined DMV and CCM \citep{klein2002generative} model, where each tree is scored with the product of the probabilities from the individual models, outperform either individual model. These results demonstrate that the two varieties of unsupervised parsing models can benefit from ensembling. In contrast, our model considers both phrase- and dependency structure jointly. \citet{seginer2007fast} introduces a parser using a representation like dependency structure, which helps constituency parsing.

\citet{bikel2004intricacies}'s analysis of prominent models at the time found that lexical dependencies provided only very minor benefits and that choosing appropriate smoothing parameters was key to performance and robustness.
\citet{Hockenmaier2002} also explores this for combinatorial categorial grammar (CCG), showing that lexical sparsity and smoothing have dramatic effects regardless of the formalism. The sparsity and expense of lexicalized PCFGs have precluded their use in most contexts, though \cite{prescher-2005-head} proposes a latent-head model to alleviate the sparse data problem.

\section{Conclusion}
\label{sec:conclusion}
In this paper, we propose neural L-PCFG, a neural parameterization method for lexicalized PCFGs, for both unsupervised dependency parsing and constituency parsing. We also provide a variational inference method to train our model. By modeling both representations together, our approach outperforms methods specially designed for either grammar formalism alone.  

Importantly, our work also adds novel insights for the unsupervised grammar induction literature by probing the role that factorizations and initialization have on model performance.  Different factorizations of the same probability distribution can lead to dramatically different performance and should be viewed as playing an important role in the inductive bias of learning syntax.  Additionally, where others have used pretrained word vectors before, we show that they too contain abstract syntactic information which can bias learning.

Finally, while out of scope for one paper, our results point to several interesting potential roads forward, including the study of the effectiveness of jointly modeling constituency-dependency representations on freer word order languages, and whether other distributed word presentations (e.g. large-scale transformers) might provide even stronger syntactic signals for grammar induction. 

Despite the demonstrated success of lexical dependencies, it should be noted that these are only unilexical dependencies, in contrast to bilexical dependencies, which also consider the dependencies between head and dependent words. Modeling these dependencies would require marginalizing over all possible dependents for each span-head pair. In this case, the time complexity of exhaustive dynamic programming over one sentence would become $\mathcal{O}(L^5|\mathcal{N}|(|\mathcal{N}| + |\mathcal{P}|)^2)$, where $L$ stands for the length of the sentence.
Assuming enough parallel workers, this time complexity can be reduced to $\mathcal{O}(L)$, but it still requires $\mathcal{O}(L^4|\mathcal{N}|(|\mathcal{N}| + |\mathcal{P}|)^2)$ auxiliary space. In contrast, our model runs for $\mathcal{O}(L^4|\mathcal{N}|(|\mathcal{N}| + |\mathcal{P}|)^2)$.
Assuming enough parallel workers, this time complexity can also be reduced to $\mathcal{O}(L)$, but still requires $\mathcal{O}(L^3|\mathcal{N}|(|\mathcal{N}| + |\mathcal{P}|)^2)$ auxiliary space. These auxiliary data can be stored in a 32GB graphic card in our experiments (e.g. with $N=20$), while the bilexical model cannot.
There are several potential methods to side-step this problem, including the use of sampling in lieu of dynamic programming, using heuristic methods to prune the grammar, and designing acceleration methods on GPU \cite{hall-etal-2014-sparser}.

\section*{Acknowledgements}

This work was supported by the DARPA GAILA project (award HR00111990063), and some experiments made use of computation credits graciously provided by Amazon AWS.
The views
and conclusions contained in this document are
those of the authors and should not be interpreted as representing the official policies, either
expressed or implied, of the U.S. Government.
The U.S. Government is authorized to reproduce
and distribute reprints for Government purposes
notwithstanding any copyright notation here on.
The authors would like to thank Junxian He and Yoon Kim for helpful feedback about the project. 

\iftaclpubformat
\fi
\bibliography{tacl2018}
\bibliographystyle{acl_natbib}

\end{document}